\pdfoutput=1

\documentclass[11pt]{article}

\usepackage[]{ACL2023}

\usepackage{times}
\usepackage{latexsym}

\usepackage[T1]{fontenc}

\usepackage[utf8]{inputenc}

\usepackage{microtype}

\usepackage{inconsolata}

\usepackage{adjustbox}
\usepackage{booktabs}
\usepackage{graphicx}
\usepackage{siunitx}
\usepackage{enumitem}

\newcommand{\pbexamples}{PB-Examples}
\newcommand{\pbunseen}{PB-Unseen}
\newcommand{\challenge}{Challenge-SRL}
\newcommand{\semlink}{Parallel-SemLink}

\title{Exploring Non-Verbal Predicates in Semantic Role Labeling:\\Challenges and Opportunities}

\author{Riccardo Orlando $^*$ \qquad Simone Conia $^*$ \qquad Roberto Navigli \\
         Sapienza NLP Group, Sapienza University of Rome \\
         \texttt{\{orlando,navigli\}@diag.uniroma1.it} \\
         \texttt{conia@di.uniroma1.it}}

\begin{document}
\maketitle

\begin{abstract}
Although we have witnessed impressive progress in Semantic Role Labeling (SRL), most of the research in the area is carried out assuming that the majority of predicates are verbs.
Conversely, predicates can also be expressed using other parts of speech, e.g., nouns and adjectives.
However, non-verbal predicates appear in the benchmarks we commonly use to measure progress in SRL less frequently than in some real-world settings -- newspaper headlines, dialogues, and tweets, among others.
In this paper, we put forward a new PropBank dataset which boasts wide coverage of multiple predicate types. Thanks to it, we demonstrate empirically that standard benchmarks do not provide an accurate picture of the current situation in SRL and that state-of-the-art systems are still incapable of transferring knowledge across different predicate types.
Having observed these issues, we also present a novel, manually-annotated challenge set designed to give equal importance to verbal, nominal, and adjectival predicate-argument structures. We use such dataset to investigate whether we can leverage different linguistic resources to promote knowledge transfer.
In conclusion, we claim that SRL is far from ``solved'', and its integration with other semantic tasks might enable significant improvements in the future, especially for the long tail of non-verbal predicates, thereby facilitating further research on SRL for non-verbal predicates.
We release our software and datasets at \url{https://github.com/sapienzanlp/exploring-srl}.
\end{abstract}

\def\thefootnote{*}\footnotetext{Equal contribution.}\def\thefootnote{\arabic{footnote}}

\section{Introduction}
\label{ref:introduction}

Over the years, Semantic Role Labeling~\citep[SRL]{gildea-jurafsky-2002-srl} -- the task of identifying the semantic relations between predicates and their arguments -- has attracted continued interest.
Enticed by the prospect of acquiring one of the ingredients that might enable Natural Language Understanding \citep{navigli-etal-2022-tour}, 
the research community has striven to overcome numerous challenges in SRL. 
As a consequence, not only have automatic systems achieved impressive results on complex benchmarks~\citep{shi_simple_2019,conia_unifying_2021}, such as CoNLL-2005~\citep{carreras_introduction_2005}, CoNLL-2008~\citep{surdeanu_conll_2008}, CoNLL-2009~\citep{hajic_conll-2009_2009}, and CoNLL-2012~\citep{pradhan_conll-2012_2012}, but SRL has also been successfully leveraged to benefit a wide array of downstream tasks in Natural Language Processing and also Computer Vision, including Machine Translation~\citep{marcheggiani_exploiting_2018,raganato_mucow_2019,song_semantic_2019}, Summarization~\citep{hardy_guided_2018,liao_abstract_2018}, Situation Recognition~\citep{yatskar_situation_2016}, and Video Understanding~\citep{sadhu_visual_2021}, among others.

Notwithstanding the achievements of previous work, we argue that there is still much to be done before the research community can claim SRL is even close to being ``solved''.
One of the simplest yet erroneous assumptions about SRL is that all predicates -- or at least the majority of them -- are verbs.
Quite the contrary, predicates often manifest themselves as nouns, adjectives, and adverbs.
For example, in the sentence ``Sensational robbery at the bank during the night: two suspects on the loose!'', the word \textit{robbery} is a predicate, as it denotes an action, and its arguments are \textit{sensational} (attribute of the robbery), \textit{at the bank} (location), \textit{during the night} (time), and \textit{two suspects} (agents).
We highlight two potential issues in the above example.
First, an SRL system that analyzes only verbal predicates cannot identify the nominal event in the sentence and, in turn, its semantic constituents.
Second, nominal events like those expressed in the above sentence are far from rare, being commonly found in several settings, such as newspaper headlines, blog titles, short messages, tweets, and dialogues.

Perhaps surprisingly, there is limited work on non-verbal predicates, mostly focused on transferring ``knowledge'' about verbal predicates to nominal ones~\citep{zhao_unsupervised_2020,klein_qanom_2020}.
The scarcity of studies on non-verbal predicates might be explained by the way in which current datasets for SRL are designed, as they focus primarily on verbal predicates~\citep{daza-frank-2020-xsrl,tripodi_united-srl_2021,jindal-etal-2022-universal}.
Therefore, any progress on non-verbal predicates is often overshadowed by the predominance of verbal instances, resulting in an incomplete picture of the actual situation.
The issue is also exacerbated by the fact that, oftentimes, benchmark results are taken at face value.
Instead, carrying out in-depth analyses is fundamental, as neural networks have been found to learn patterns that are different from those of humans, especially in semantic tasks~\citep{maru_nibbling_2022}. 
In this paper, we perform a reality check and explore non-verbal predicates in English SRL.
More specifically, our contributions are as follows:
\begin{itemize}[itemsep=5pt]
    \item We provide an empirical demonstration that state-of-the-art systems are not capable of generalizing from verbal to nominal and adjectival predicate-argument structures (PAS) in PropBank-based SRL;

    \item We investigate whether other PAS inventories -- namely, FrameNet, VerbNet, and VerbAtlas -- are better suited for transferring learned patterns across predicate types;

    \item We introduce a novel, manually-annotated challenge set to evaluate current and future SRL systems on verbal, nominal, and adjectival PAS;

    \item We analyze possible directions and strategies for prospective work on non-verbal SRL.
\end{itemize}

\section{Challenges}
As mentioned above, relying on standard benchmarks does not allow us to properly evaluate the performance of state-of-the-art systems on non-verbal SRL.
Cases in point are the CoNLL Shared Tasks: CoNLL-2005 covers only verbal predicates; CoNLL-2009 includes verbal and nominal predicates but makes it difficult to compare them, as they belong to two different inventories, PropBank and NomBank, respectively; CoNLL-2012 and its revision in OntoNotes 5.0~\cite{pradhan-etal-2022-propbank} do not cover adjectival predicates.
Therefore, identifying unaddressed challenges, especially in non-verbal SRL, is far from trivial.

\begin{table}[t]
    \centering
    \begin{adjustbox}{width=0.95\linewidth}
    \begin{tabular}{lcccc}
        \toprule
        
       & \textbf{Verbs} & \textbf{Nouns} & \textbf{Adjs} & \textbf{Framesets} \\

        \cmidrule(r){2-4} \cmidrule(l){5-5}

        
        CoNLL-2009 & 1090 & 1337 & ~~~~~~0 & 2427 \\
        OntoNotes 5.0 & 2215 & ~~782 & ~~~~~~3 & 2490 \\
        \pbexamples{} & 5465 & 1384 & 1599 & 7481 \\
        \pbunseen{} & 2457 & ~~469 & 1389 & 4001 \\

        \bottomrule
    \end{tabular}
    \end{adjustbox}
    \caption{Comparison of the coverage of each evaluation benchmark in terms of unique framesets by part of speech (i.e. according to their association with predicate occurrences from the various parts of speech), and total number of part-of-speech independent framesets. \pbexamples{} and \pbunseen{} provide more extensive coverage than CoNLL-2009 and OntoNotes 5.0.}
    \label{tab:pb-examples-stats}
\end{table}

\begin{table*}[t]
    \centering
    \begin{adjustbox}{max width=\linewidth}
    \begin{tabular}{lccccccccccccc}
        \toprule
         & \multicolumn{3}{c}{\textbf{OntoNotes}} && \multicolumn{4}{c}{\textbf{\pbexamples{}}} && \multicolumn{4}{c}{\textbf{\pbunseen{}}} \\
         
         \cmidrule{2-4} \cmidrule{6-9} \cmidrule{11-14}
         
         & \textbf{Verbs} & \textbf{Nouns} & \textbf{V+N} && \textbf{Verbs} & \textbf{Nouns} & \textbf{Adjs} & \textbf{V+N+A} && \textbf{Verbs} & \textbf{Nouns} & \textbf{Adjs} & \textbf{V+N+A} \\

        \midrule

        \textit{Predicates} \\
        CN-22$_\textbf{\small\ verbs}$ & 95.4 & 83.5 & 94.1 && 79.1 & 70.7 & 54.0 & 74.7 && 46.8 & 34.3 & 42.8 & 51.4 \\ 
        CN-22$_\textbf{\small\ nouns}$ & 47.6 & 96.5 & 53.4 && 65.6 & 75.4 & 59.5 & 69.7 && 62.0 & 60.3 & 54.9 & 64.1 \\
        CN-22$_\textbf{\small\ verbs + nouns}$ & 95.4 & 96.5 & 95.6 && 80.7 & 80.0 & 56.4 & 77.5 && 51.1 & 38.5 & 45.1 & 53.6 \\

        \midrule

        \textit{Roles} \\
        CN-22$_\textbf{\small\ verbs}$ & 84.7 & 16.4 & 80.2 && 57.8 & 34.6 & 25.1 & 49.6 && 25.6 & ~~6.8 & 16.5 & 26.1 \\
        CN-22$_\textbf{\small\ nouns}$ & 11.2 & 72.8 & 16.2 && 15.1 & 45.1 & ~~5.4 & 22.1 && 15.4 & 29.1 & ~~4.2 & 16.3 \\
        CN-22$_\textbf{\small\ verbs + nouns}$ & 84.7 & 76.1 & 84.1 && 59.7 & 59.1 & 25.6 & 55.2 && 28.9 & 17.8 & 16.7 & 28.5 \\
        
        \bottomrule
    \end{tabular}
    \end{adjustbox}
    \caption{F1 scores of CN-22 on the test sets of OntoNotes, \pbexamples{}, and \pbunseen{}, divided by predicate type. These results show that a state-of-the-art system is not capable of ``transferring knowledge'' from one predicate type to another, e.g., from verbs to nouns or vice versa.}
    \label{tab:cross-type-results}
\end{table*}

\paragraph{Introducing \pbexamples{} and \pbunseen{}.}
Since OntoNotes 5.0 -- the largest gold evaluation framework for PropBank-based SRL -- does not comprehensively evaluate different predicate types, we collect the example sentences provided with each predicate in PropBank 3~\citep{palmer_proposition_2005,pradhan-etal-2022-propbank} to create a new evaluation benchmark, named \pbexamples{}.
This allows us to build a ``controlled'' benchmark, the first on which we can evaluate the performance of PropBank-based SRL on verbal, nominal, and adjectival PAS.

In Table~\ref{tab:pb-examples-stats} we report statistics on the coverage of CoNLL-2009, OntoNotes 5.0 and \pbexamples{} in terms of unique framesets (rightmost column), where the considerably higher frameset coverage of \pbexamples{} is evident.
Compared to its alternatives, \pbexamples{} covers 7481 unique PropBank framesets against 2490 framesets covered in the OntoNotes test set and 2427 in CoNLL-2009.
Moreover, when comparing \pbexamples{} to OntoNotes, the number of unique framesets used in verbal predicate occurrences is more than double (5465 vs. 2215), whereas it is almost double for nominal occurrences (1384 vs. 782).
Adjectival occurrences are essentially missing in OntoNotes (with 3 unique framesets only), while \pbexamples{} covers 1599.
We remark that the same PropBank frameset can be used to annotate predicate occurrences from different parts of speech, which explains why the total number of unique framesets does not correspond to the sum of framesets used for verbal, nominal and adjectival predicate occurrences (second, third and fourth column of Table~\ref{tab:pb-examples-stats}).

Given its considerably higher coverage, \pbexamples{} also enables a solid evaluation of an SRL system on over 4000 predicate senses that are not included in OntoNotes 5.0; we call this more challenging testbed \pbunseen{}.
We report statistics on \pbunseen{} in the last row of Table~\ref{tab:pb-examples-stats}.

\paragraph{Cross-type knowledge transfer.}
Now that we have wide-coverage multi-type SRL datasets, we can test the ability of SRL systems to generalize across types. The main objective of our experiments here is to empirically demonstrate that: i) ``knowledge transfer'' between predicate types is an unaddressed challenge, and ii) this problem is not apparent in OntoNotes, but becomes evident from \pbexamples{} and \pbunseen{}.
To prove these points, we take CN-22 -- a state-of-the-art system~\cite{conia_probing_2022} -- and study its behavior when trained on the entire OntoNotes (CN-22$_\textbf{\small verbs+nouns}$), only on its verbal structures (CN-22$_\textbf{\small verbs}$), or only on its nominal structures (CN-22$_\textbf{\small nouns}$).
The results on the test set of OntoNotes, shown in Table~\ref{tab:cross-type-results}, represent the first evidence that even a state-of-the-art SRL system is affected by limited generalization capabilities across predicate types. Indeed, the performance of CN-22$_\textbf{\small verbs}$ drops significantly when evaluated on nominal PAS, from 84.7 to 16.4 points in F1 score on argument labeling, and that of CN-22$_\textbf{\small nouns}$ drops analogously when evaluated on verbal instances, from 72.8 to 11.2 on argument labeling.

One could observe that CN-22$_\textbf{\small verbs+nouns}$, jointly trained on verbal and nominal instances, seems to solve the cross-type transfer problem.
However, this is true only because the OntoNotes test set does not feature adjectival structures.
Indeed, it is very clear from the results on our \pbexamples{} and \pbunseen{} that the performance of CN-22$_\textbf{\small verbs+nouns}$ does not improve on adjectival PAS compared to CN-22$_\textbf{\small verbs}$ (only +0.5\% on \pbexamples{} and +0.2\% on \pbunseen{} for argument labeling).
Therefore, we can derive that joint learning on two predicate types (i.e. the verbal and nominal ones) does not provide breakthrough improvements on a third predicate type (i.e. the adjectival one).
We stress that, in this case, we cannot simply rely on jointly training CN-22 on verbal, nominal, and adjectival instances as, to our knowledge, no training dataset includes adjectival PAS for PropBank-based SRL.

\section{Opportunities}

In the previous Section, our experiments show that zero-shot knowledge transfer across predicate types is still challenging.
We argue that this problem is caused by two main factors.
First, PropBank was not designed to aid cross-type knowledge transfer, e.g., the nominal predicate \textit{theft.01} is not linked to its verbal equivalent \textit{steal.01}.
Second, recent SRL systems might have limited capability for recognizing common patterns across different predicate types.
We conduct an initial investigation of these aspects and discuss some opportunities for improving non-verbal SRL.

\paragraph{The role of the linguistic resource.}
While PropBank might not be the ideal resource for non-verbal SRL, other inventories -- based on different linguistic theories -- may provide features that could be helpful to aid knowledge transfer between predicate types.
After all, previous studies have already shown that language models leverage different hidden layers depending on the linguistic resource used for SRL~\cite{kuznetsov_matter_2020,conia_probing_2022}.
Here, instead, we take the opportunity to study if there is an inventory whose theoretical principles can aid the generalization capability of an existing SRL system on unseen patterns.

\begin{table}[t]
    \centering
    \begin{adjustbox}{max width=\linewidth}
    \begin{tabular}{lccccccccccccc}
        \toprule
        & \multicolumn{3}{c}{\textbf{Predicates}} && \multicolumn{3}{c}{\textbf{Roles}} \\
         \cmidrule{2-4} \cmidrule{6-8}

         & \textbf{P} & \textbf{R} &\textbf{F1} && \textbf{P} & \textbf{R} &\textbf{F1} \\

        \midrule
 
        CN-22$_\textbf{\small\ PropBank}$ & 99.1 & 96.7 & 97.9 && 88.3 & 88.0 & 88.1 \\
        CN-22$_\textbf{\small\ FrameNet}$ & 99.1 & 96.7 & 97.9  && 89.3 & 89.5 & 89.4 \\
        CN-22$_\textbf{\small\ VerbNet}$ & \textbf{99.9} & 97.4 & 98.6 && \textbf{89.8} & 89.3 & 89.5 \\
        CN-22$_\textbf{\small\ VerbAtlas}$ & 99.7 & \textbf{97.7} & \textbf{98.7} && 89.4 & \textbf{90.0} & \textbf{89.7} \\
   
        \bottomrule

    \end{tabular}
    \end{adjustbox}
    \caption{Precision (P), Recall (R), and F1 scores of CN-22 on \semlink{}. 
    For each row, we evaluate the performance of the system when trained using the related inventory, e.g., CN-22$_\textbf{\small\ PropBank}$ is trained on \semlink{} annotated with PropBank and the results are reported against the test set for the same inventory. }
    \label{tab:semlink-test-results}
\end{table}

We thus evaluate empirically the differences between four different inventories, namely, PropBank, FrameNet \cite{framenet}, VerbNet \cite{Schuler2005VerbnetAB}, and VerbAtlas \cite{di_fabio_verbatlas_2019}.\footnote{Appendix \ref{sec:appendix-intentories} provides an overview of the inventories.}
To do this, we create \semlink{}, a multi-inventory benchmark made up of the subset of OntoNotes from SemLink 2.0~\cite{stowe_semlink_2021}, whose predicates and arguments are annotated with PropBank, FrameNet, and VerbNet.
We also include VerbAtlas annotations thanks to the inter-resource mapping between VerbNet, WordNet, and VerbAtlas.\footnote{Appendix \ref{sec:appendix-mapping} provides further details on our mapping procedure.}
For each of these inventories, \semlink{} includes a training, a validation, and a test set with 7336, 816, and 906 sentences, respectively.

While we stress that this experimental setting is severely limited since it assumes that all resources can be mapped to each other 1-to-1, it provides a controlled environment for a fair, direct comparison.
To study the impact of the inventory, we evaluate our SRL system on each of the linguistic inventories in \semlink{} (CN-22$_\textbf{\small\ PropBank}$, CN-22$_\textbf{\small\ FrameNet}$, CN-22$_\textbf{\small\ VerbNet}$, and CN-22$_\textbf{\small\ VerbAtlas}$).
The results in Table~\ref{tab:semlink-test-results} testify that the linguistic resource of choice plays a role in the results.
In particular, we can observe a relative error rate reduction of 38\% in predicate sense disambiguation (from 97.9 to 98.7) and 13\% in argument labeling (from 88.1 to 89.7) when using VerbAtlas instead of PropBank.
This result indicates that higher-level semantic abstractions, such as semantics-based clusters, as available in VerbAtlas thanks to its organization of frames as verbal synset groupings, and cross-predicate role semantics, as adopted in VerbNet and also VerbAtlas, can help a system generalize better on unseen patterns.

\begin{table}[t]
    \centering
    \begin{adjustbox}{max width=\linewidth}
    \begin{tabular}{lcccc}
        \toprule

         & \textbf{Verbs} & \textbf{Nouns} & \textbf{Adjs} & \textbf{V+N+A}\\

        \midrule

        \textit{Predicates} \\
        CN-22$_\textbf{\small\ PropBank}$ & 14.5 & 22.2 & 27.7 & 21.7 \\ 
        CN-22$_\textbf{\small\ VerbAtlas}$ & 49.4 & 17.7 & 13.5 & 26.0 \\

        \midrule

        \textit{Roles} \\
        CN-22$_\textbf{\small\ PropBank}$ & ~~5.5 & ~~2.1 & 10.8 & 54.2 \\
        CN-22$_\textbf{\small\ VerbAtlas}$ & 47.0 & 44.2 & 36.8 & 42.8 \\

        \bottomrule

    \end{tabular}
    \end{adjustbox}
    \caption{F1 scores of CN-22 on \challenge{}.}
    \label{tab:hand-test-results}
\end{table}

\paragraph{\challenge{}.}
While our multi-inventory SemLink-based dataset provides a preliminary indication of the role of a linguistic inventory, it only includes verbal predicates.
To further validate the preliminary results obtained on our multi-inventory SemLink-based dataset, we create a small challenge test set for verbal, nominal, and adjectival SRL, manually annotated with parallel labels for PropBank, the most popular inventory, and VerbAtlas, the most promising inventory (cf. Table \ref{tab:semlink-test-results}).
This new test set is particularly challenging, as it features only PAS that do not appear in OntoNotes.
Therefore, \challenge{} makes it possible to measure the capability of an SRL system to generalize i) across predicate types, and ii) on the long tail of predicate senses.

To construct \challenge{}, we randomly selected a total of 288 sentences -- 96 sentences for each predicate type -- from \pbunseen{}.
We then asked three expert annotators to independently annotate each sentence with predicate senses and their semantic roles.
The annotation process was carried out in two phases: first, each person annotated each sentence independently, resulting in a disagreement of 32\%; then, the annotators discussed and resolved their disagreements, if possible, reducing them to 6\%.
Overall, \challenge{} includes 1898 predicate-argument pairs.

As we can see from Table~\ref{tab:hand-test-results}, \challenge{} confirms our preliminary experiments, macroscopically magnifying the differences between PropBank and VerbAtlas.
First, we observe that VerbAtlas is significantly better in predicate sense disambiguation for verbal instances (49.5 vs. 14.5 in F1 score) but worse for nominal and adjectival ones (22.2 vs. 17.7 and 27.7 vs. 13.5, respectively).
This is mainly because VerbAtlas was not designed for non-verbal SRL and, therefore, it does not provide a lemma-to-sense dictionary to restrict the possible frames of nominal and adjectival predicates.
Second, VerbAtlas significantly outperforms PropBank on argument labeling of verbs (47.0 vs. 5.5 in F1 score), nouns (44.2 vs. 2.1), and adjectives (36.8 vs. 10.8).
We argue that this is largely due to the adoption in VerbAtlas of cross-frame semantic roles that are coherent across frames, 
which allows the system to leverage other predicates seen at training time with similar structures.

\paragraph{Leveraging Word Sense Disambiguation.}

\begin{table}[t]
    \centering
    \begin{adjustbox}{max width=\linewidth}
    \begin{tabular}{lcccc}
        \toprule
         
         & \textbf{Verbs} & \textbf{Nouns} & \textbf{Adjs} & \textbf{V+N+A} \\

        \midrule

        CN-22$_\textbf{\footnotesize\ SemLink}$ & ~~6.2 & ~~6.2 & ~~3.1 & ~~5.2 \\ 
        CN-22$_\textbf{\footnotesize\ OntoNotes}$ & 49.4 & ~~5.2 & 10.2 & 26.0 \\
        WSD$_\textbf{\footnotesize\ baseline}$ & 46.7 & 32.7 & ~~3.8 & 31.7 \\
        Oracle$_\textbf{\footnotesize\ SL+WSD}$ & 58.9 & 37.2 & ~~9.3 & 31.4 \\
        Oracle$_\textbf{\footnotesize\ ON+WSD}$ & \textbf{60.5} & \textbf{41.6} & \textbf{25.6} & \textbf{41.5} \\
        \bottomrule
         
    \end{tabular}
    \end{adjustbox}
    \caption{F1 scores on predicate disambiguation of CN-22 and a WSD system on \challenge{}. The scores of Oracle$_\textbf{\small\ SL+WSD}$ (Oracle$_\textbf{\small\ ON+WSD}$) are obtained by picking the best prediction between WSD$_\textbf{\small\ baseline}$ and CN-22$_\textbf{\small\ SemLink}$ (CN-22$_\textbf{\small\ OntoNotes}$).}
    \label{tab:wsd-comparison}
\end{table}

Finally, we carry out a preliminary exploration of possible directions that could aid non-verbal SRL in the future.
While SRL research has not dealt with non-verbal semantics, other areas have investigated semantics for different parts of speech, and one of these is Word Sense Disambiguation (WSD).
More specifically, WSD is the task of assigning the most appropriate sense to a word in context according to a predefined sense inventory \cite{bevilacqua-etal-2021-recent}.
It is easy to notice how this task resembles predicate sense disambiguation in SRL, the only difference being that WSD is not limited to predicates, as it aims to disambiguate every content word.
Therefore, we believe that WSD is an interesting candidate to explore whether a different disambiguation task can help to improve the generalization capability of an existing SRL system on \challenge{}, i.e., on predicate-argument structures that the SRL system did not see at training time.

To investigate the effect of WSD on SRL, we start by leveraging the fact that VerbAtlas frames are clusters of WordNet synsets. 
Therefore, we map each synset predicted by AMuSE-WSD \cite{orlando_amuse-wsd_2021,orlando-etal-2022-universal},\footnote{\url{https://nlp.uniroma1.it/amuse-wsd/}} a state-of-the-art off-the-shelf WSD system, to a VerbAtlas frame, and compare them to the prediction of our SRL system.
Table~\ref{tab:wsd-comparison} shows the performance of AMuSE-WSD on predicate sense disambiguation (WSD$_\textrm{baseline}$).
Interestingly, we observe that a simple WSD baseline can strongly outperform an SRL system when training data is scarce.
Indeed, AMuSE-WSD surpasses CN-22$_\textbf{\small\ SemLink}$ in each predicate type (46.7 vs 6.2, 32.7 vs 6.2, 3.8 vs 3.1, for verbs, nouns and adjectives,  respectively), and CN-22$_\textbf{\small\ OntoNotes}$ in nominal predicates, with an overall improvement of +5.7 (31.7 vs 26.0) over the best performing SRL system.

Most interestingly, if we employ an oracle to pick the best prediction between the WSD baseline and our best SRL system, we notice a further improvement (41.5\% vs. 26.0\%), demonstrating that current state-of-the-art SRL systems can still benefit from explicit lexical semantics.
We hypothesize that tighter integration of the two tasks may lead to even better improvements in generalization capabilities.

\section{Conclusion and Future Work}
In this paper, we carried out a reality check and demonstrated that, despite impressive results on standard benchmarks by state-of-the-art systems, SRL is still far from ``solved''.
Indeed, thanks to a carefully-designed set of experiments and the introduction of novel, manually-curated, wide-coverage benchmarks, we showed that current SRL systems possess inadequate capabilities for transferring knowledge between predicate types.

Our analyses pointed out that we can address this limitation by working in two directions: leveraging the intrinsic characteristic of frameset resources, including semantics-based clusters and cross-predicate role semantics, and tighter integration of other semantics-based tasks, such as Word Sense Disambiguation, into SRL.

We hope our work will be a stepping stone for innovative research on high-performance SRL systems for non-verbal predicate-argument structures, a problem that still needs extensive investigation.
For this reason, we release our software and datasets at \url{https://github.com/sapienzanlp/exploring-srl}.

\section*{Limitations}
Part of our analyses and experiments is based on our \semlink{} dataset, which provides parallel annotations for PropBank, FrameNet, VerbNet, and VerbAtlas.
We take the opportunity to remark that this is a constrained setting, as these resources cannot be mapped 1-to-1 without losing information.
As such, this setting may not provide the full picture of how these resources compare against each other.
However, we also believe that a setting like this can at least provide an intuitive idea of the role of a linguistic resource in cross-inventory generalization.
Creating novel benchmarks that can better compare the role of different linguistic resources is certainly a direction for future work that may provide novel insights into verbal and non-verbal SRL.

Another limitation of our work is the small size of \challenge{}.
Even though \challenge{} contains only about 300 sentences, it features almost 2000 predicate-argument pairs, and this is a number that is sufficient to show the inability of a current state-of-the-art system to generalize across predicate types.
We acknowledge that a larger benchmark may have provided further insights.
However, we also note that, in our case, increasing the number of annotations would hardly have brought us to a different conclusion, especially given the large differences in performance among the model configurations that we evaluated.

Finally, we stress that our experiments on integrating a simple WSD baseline into an SRL system do not provide a definitive answer on whether more complex integrations may lead to improved results.
Instead, our intention is to support the claim that SRL is still far from being ``solved'', as knowledge from other tasks can still hypothetically bring benefits to an existing SRL system, especially when the size of the training data is small.

\section*{Ethics Statement}
We release all the new datasets we produce under an open license.
However, some of the datasets mentioned and used in our paper are not openly available, e.g., CoNLL-2009 and OntoNotes 5.0.
We acknowledge the fact that such datasets may become unavailable at a later moment, as their distribution is not under our control.


\section*{Acknowledgments}
\begin{center}
\noindent
    \begin{minipage}{0.1\linewidth}
        \begin{center}
            \includegraphics[scale=0.2]{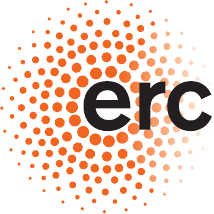}
        \end{center}
    \end{minipage}
    \hspace{0.01\linewidth}
    \begin{minipage}{0.70\linewidth}
        The authors gratefully acknowledge the support of the ERC Consolidator Grant MOUSSE No.\ 726487 under the European Union's Horizon 2020 research and innovation programme.
    \end{minipage}
    \hspace{0.01\linewidth}
    \begin{minipage}{0.1\linewidth}
        \begin{center}
            \includegraphics[scale=0.07]{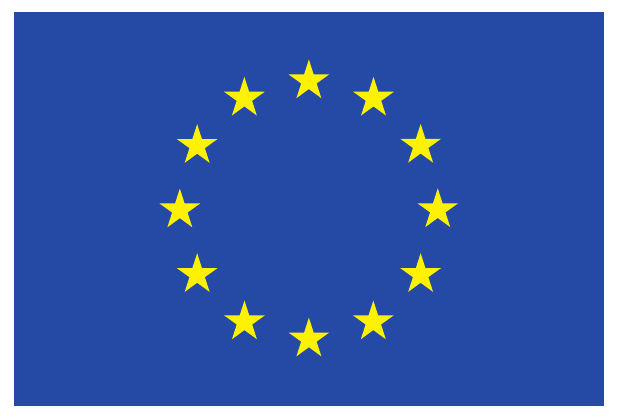}
        \end{center}
    \end{minipage}\\
\end{center}

The last author gratefully acknowledges the support of the PNRR MUR project PE0000013-FAIR.

\bibliography{main}
\bibliographystyle{acl_natbib}

\appendix

\section{Inventories}
\label{sec:appendix-intentories}

In this paper, we evaluate empirically how SRL systems are influenced by the different linguistic inventories employed. We tested four popular inventories, namely PropBank, FrameNet, VerbNet, and VerbAtlas. Each of these inventories features different characteristics, which we summarize briefly here.

\paragraph{PropBank} PropBank \cite{palmer_proposition_2005} enumerates the senses of each predicate lemma, e.g., \textit{eat.01}, \textit{eat.02}, etc., and defines semantic roles (\textsc{\small Arg0}-\textsc{\small Arg5}) that are specific to each predicate sense, e.g., the meaning of \textsc{\small Arg2} in \textit{eat.01} differs from that of \textit{eat.02}.

\paragraph{FrameNet} FrameNet \cite{framenet} groups predicates that evoke similar actions in semantic frames, e.g., the frame \textit{Ingestion} includes eating, feeding, devouring, among others; each frame can have frame-specific roles, e.g., \textsc{\small Ingestor} and \textsc{\small Ingestible}.

\paragraph{VerbNet} VerbNet \cite{Schuler2005VerbnetAB} defines classes of verbs with similar syntactic patterns, e.g., eating and drinking belong to \textit{Eat-39.1-1}; all verb classes share a set of thematic roles, e.g., \textsc{\small Agent} and \textsc{\small Patient}.

\paragraph{VerbAtlas} VerbAtlas \cite{di_fabio_verbatlas_2019} clusters WordNet \cite{miller_wordnet_1992} synsets into coarse-grained frames, similar to FrameNet, and adopts a common set of thematic roles for all frames, similar to VerbNet.

\section{\semlink{}}
In this Section, we provide further details on the construction process of \semlink{}.
We leverage the data distributed as part of SemLink 2.0 \cite{stowe_semlink_2021}, which includes instances from OntoNotes 5.0 annotated with PropBank, FrameNet, and VerbNet. 
We select the subset of the instances that have a corresponding annotation in all three inventories.
In addition, we also include VerbAtlas annotations through the inter-resource mapping between VerbNet, WordNet, and VerbAtlas. 
To convert the predicate senses, we employ the mapping from VerbNet to WordNet included in the Unified Verb Index (UVI)\footnote{\url{https://uvi.colorado.edu/}} project: since a VerbAtlas frame is a cluster of WordNet synsets, we associate a VerbNet class with a VerbAtlas frame through their corresponding synset.
Additionally, we also extend the VerbAtlas annotations to include argument roles. 
Given that both VerbNet and VerbAtlas adopt a similar set of thematic roles, we manually map all the VerbNet roles to their corresponding VerbAtlas ones and convert the argument annotations accordingly.

\section{Mapping Nouns to VerbAtlas Frames}
\label{sec:appendix-mapping}
Since VerbAtlas was originally designed only as a verbal inventory, its frames contain only verbal WordNet synsets.
To expand its coverage and include nominal predicates, we propose a method for deriving nominal predicates from the verbal ones already included.
The method leverages WordNet \cite{miller_wordnet_1992}, a lexical database that contains a wealth of information about word senses and their relationships.
Specifically, we use the ``hypernym'' and ``derivationally related forms'' relations in WordNet to identify nominal word senses that are semantically related to a verbal predicate in VerbAtlas.
Informally, to be included in our expanded version of VerbAtlas, a nominal word sense must meet the following criteria:
\begin{enumerate}
    \item It must have a ``hypernym'' that belongs to the top-100 most frequent nominal senses related to \textit{event.n.01}, i.e., event as in ``something that happens at a given place and time''.
    \item It must be semantically related -- ``derivationally related forms'' related -- to a verbal predicate included in a VerbAtlas frame.
\end{enumerate}
This approach allows us to identify a large number of nominal word senses that are semantically related to a verbal predicate in VerbAtlas.
Therefore, we assign these nominal word senses to the same VerbAtlas frame as their related verbal predicates.
In total, we are able to cluster 5334 nominal word senses, significantly expanding the coverage of VerbAtlas to include both verbal and nominal predicates.
We release this mapping together with the rest of our software and datasets.

\section{Mapping Adjectives to VerbAtlas Frames}

We follow a similar strategy to also include adjectival predicates in VerbAtlas.
This time, we rely on the ``pertainyms'', ``similar to'', and ``derivationally related forms'' relations to connect adjectival word senses in WordNet to VerbAtlas frames.
More specifically, we include each adjectival word sense that satisfies at least one of the following conditions:
\begin{itemize}
    \item It must be ``derivationally related'' or ``pertaining'' to a noun or verb sense that is already included in VerbAtlas;
    \item It must be ``similar to'' another word sense that is in turn ``derivationally related'' to a predicate in VerbAtlas.
\end{itemize}
We then assign these adjectival word senses to the same VerbAtlas frame as their related verbal and nominal predicates.
As a result, we are able to include 2968 adjectival predicates in VerbAtlas.
We release this mapping together with the rest of our software and datasets.

\section{License}
We release our data under the Creative Commons Attribution Share-Alike (CC-BY-SA) license.

\end{document}